\documentclass{article}

\PassOptionsToPackage{numbers,sort&compress}{natbib}

\usepackage[preprint]{neurips_2026}

\usepackage[utf8]{inputenc} % allow utf-8 input
\usepackage[T1]{fontenc}    % use 8-bit T1 fonts
\usepackage{hyperref}       % hyperlinks
\usepackage{url}            % simple URL typesetting
\usepackage{booktabs}       % professional-quality tables
\usepackage{amsfonts}       % blackboard math symbols
\usepackage{nicefrac}       % compact symbols for 1/2, etc.
\usepackage{microtype}      % microtypography
\usepackage{xcolor}         % colors
\usepackage{graphicx}
\usepackage{amsmath}

\usepackage{tabularx}
\usepackage{colortbl} % 用于给行上色
\usepackage{array} % 必须引入，用于处理垂直居中
\title{AirQualityBench: A Realistic Evaluation Benchmark for Global Air Quality Forecasting}

\author{%
\makebox[\textwidth][c]{%
\begin{tabular}{c}
Xing Xu\textsuperscript{1,2} \quad
Xu Wang\textsuperscript{1,2,*} \quad
Yudong Zhang\textsuperscript{1,2} \quad
Huilin Zhao\textsuperscript{3} \\
Zhengyang Zhou\textsuperscript{1,2} \quad
Yang Wang\textsuperscript{1,2,*} \\
\\[-0.4em]
\textsuperscript{1}University of Science and Technology of China (USTC), Hefei, Anhui, China \\
\textsuperscript{2}Suzhou Institute for Advanced Research, USTC, Suzhou, Jiangsu, China \\
\textsuperscript{3}The Hong Kong Polytechnic University, Hong Kong, China \\
\\[-0.4em]
\texttt{xuxing2025@mail.ustc.edu.cn} \\
\\[-0.4em]
\end{tabular}%
}%
}

\begin{document}

\maketitle

\begin{abstract}
Air-quality forecasting models are commonly evaluated on regional, preprocessed, and normalized datasets, where missing observations are removed or artificially completed. Such protocols simplify comparison but hide the conditions that dominate real monitoring networks: uneven global coverage, structured missingness, heterogeneous pollutant scales, and deployment cost. We introduce \textbf{AirQualityBench}, a global multi-pollutant benchmark designed to evaluate forecasting models under these realistic conditions. The benchmark contains hourly observations from 3,720 monitoring stations over 2021--2025, covers six major pollutants, and preserves provider-native observation masks. Rather than imputing a dense data tensor, AirQualityBench exposes missingness as part of the forecasting problem and reports errors on valid future observations after inverse transformation to physical concentration scales. Evaluating representative spatio-temporal models under this unified protocol shows that strong performance on sanitized datasets does not reliably transfer to global, fragmented monitoring streams. AirQualityBench therefore serves as a realistic testbed for scalable, mask-aware, and physically interpretable air-quality forecasting. All benchmark data, code, evaluation scripts, and baseline implementations are available at \href{https://github.com/Star-Learning/AirQualityBench}{GitHub}.
\end{abstract}

\section{Introduction}
\label{sec:intro}

% Monitoring and forecasting air quality have become foundational pillars of global public health governance. Unlike simple time-series data, atmospheric pollutants exhibit complex spatio-temporal dynamics characterized by long-range transport, regional diffusion, and intricate chemical transformations\cite{du2019deep,han2023machine,chen2025comprehensive}.
% Consequently, Deep Spatio-Temporal Graph (STG) Learning has emerged as the indispensable paradigm for this task. By modeling monitoring stations as nodes in a dynamic graph, these frameworks can simultaneously capture latent non-linear spatial correlations and multi-scale temporal dependencies. Such ability to decipher complex atmospheric patterns within massive observational streams offers unprecedented potential for precise environmental decision-making and early-warning systems.

Air-quality forecasting is a challenging spatio-temporal prediction problem because pollutant dynamics are shaped by regional transport, meteorology, emissions, and heterogeneous monitoring coverage.\cite{national2010global,du2019deep,world2021global,chen2025comprehensive} Recent deep spatio-temporal models have achieved strong results on curated regional datasets, but their progress depends critically on whether the underlying benchmarks reflect real monitoring conditions. The rapid evolution of these predictive architectures relies heavily on rigorous benchmarks, which serve as the "yardstick" for measuring algorithmic progress and ensuring reproducibility~\cite{liu2023largest}. High-quality air quality benchmarks are particularly vital as they provide the standardized environment necessary to validate a model's ability to generalize across diverse urban topologies and varying climatic conditions. In the field of spatio-temporal learning, established datasets have historically catalyzed breakthroughs by providing a common ground for comparing disparate neural architectures, such as Graph Convolutional Networks (GCNs) and Transformers, thereby bridging the gap between theoretical research and real-world deployment.

Despite their value, existing public air-quality forecasting benchmarks often simplify several aspects that are central to deployment-oriented evaluation. \textbf{First}, existing benchmarks~\cite{wang2020pm2,liu2021new,wang2025pcdcnet} are geographically confined to single cities or specific administrative regions with limited node scales, which
effectively severs the planetary-scale transport patterns of pollutants and prevents models from validating their scalability in large-scale networks. \textbf{Second}, these benchmarks frequently employ manual interventions to construct "sanitized" data tensors, such as linear interpolation or mean
imputation. This deviates from the physical reality of \textit{Missing Not At Random} (MNAR) patterns~\cite{vicent2026spatio}, leading to a pervasive synthetic bias that fails to reflect model robustness against fragmented, real-world observation streams. \textbf{Third}, the prevailing reliance on
unitless, normalized evaluation metrics obscures the physical interpretability and toxicological significance of prediction errors~\cite{li2023development, hunt2025stop}, creating a profound gap between laboratory experiments and the requirements of environmental regulatory deployment.

We argue that the central obstacle is not only the lack of a larger air-quality dataset, but the lack of a realistic evaluation protocol. Existing benchmarks often simplify the monitoring process into dense or partially completed tensors and report errors on normalized scales. While useful for controlled experiments, this setting removes exactly the factors that determine whether a forecasting model can be deployed in real sensing networks: missing observations, non-uniform spatial coverage, pollutant-specific scales, and computational cost. AirQualityBench is designed around these factors. It evaluates whether models can forecast from fragmented multi-pollutant histories, remain robust across uneven global monitoring networks, and produce errors that are interpretable in physical concentration units.

\begin{figure}[t]
    \centering
    %\fbox{\rule[-.5cm]{0cm}{4cm} \rule[-.5cm]{4cm}{0cm}}
    \includegraphics[width=\linewidth]{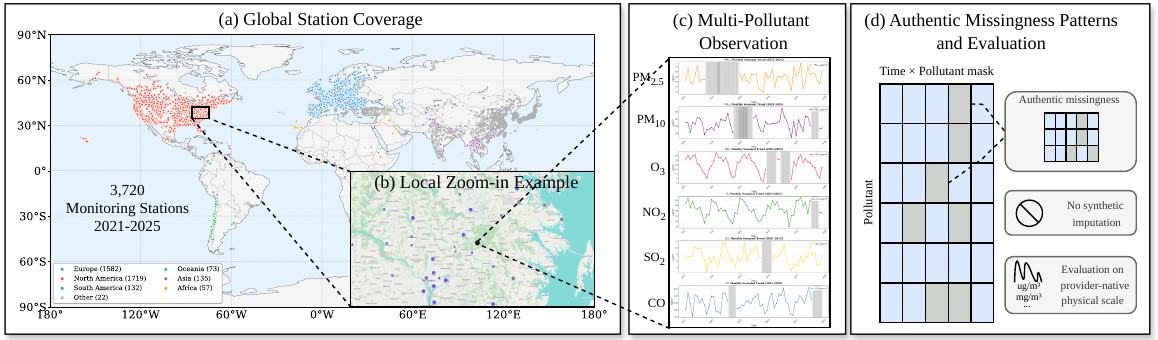} %
    \caption{\textbf{Overview of AirQualityBench.} The benchmark combines a global network of 3,720 monitoring stations, synchronized observations of six pollutants, authentic missingness patterns, and physical-scale evaluation, providing a realistic testbed for large-scale spatio-temporal air quality forecasting.}
    \label{fig:introduction}
\end{figure}

% To address these scholarly gaps and redefine the boundaries of spatio-temporal forecasting, we introduce \textbf{AirQualityBench}. Rather than offering another "sanitized" dataset, AirQualityBench serves as a standardized evaluation paradigm designed to restore physical reality to atmospheric modeling (Fig.~\ref{fig:introduction}). Our core contributions are as follows:
To support realistic and reproducible evaluation of air-quality forecasting models, we introduce \textbf{AirQualityBench}, a global multi-pollutant benchmark with authentic missingness and physical-scale reporting (Fig.~\ref{fig:introduction}). Our contributions are threefold:

\begin{itemize}
    \item \textbf{Global multi-pollutant benchmark.} 
    AirQualityBench contains hourly observations from 3,720 resident monitoring stations over 2021--2025, covering PM$_{2.5}$, PM$_{10}$, O$_3$, NO$_2$, SO$_2$, and CO. It provides a geographically diverse testbed for evaluating scalability and cross-region heterogeneity beyond regional benchmarks.

    \item \textbf{Mask-aware physical-scale evaluation.} 
    We preserve provider-native observation masks instead of constructing fully imputed tensors, and evaluate predictions only on valid future measurements. Errors are reported after inverse transformation to pollutant-specific concentration scales, enabling physically interpretable metric.

    \item \textbf{Unified baselines and diagnostics.} 
    We evaluate representative spatio-temporal forecasting models under the same split, masking protocol, graph construction, and scripts, revealing their behavior under pollutant-specific sparsity, non-uniform station coverage, and accuracy--efficiency trade-offs.
\end{itemize}

\section{Problem Definition and Related Work}
\label{sec:preliminary_related}

In this section, we formally define the masked spatio-temporal forecasting task
and review the relevant literature in deep spatio-temporal learning and air
quality modeling.

\subsection{Problem Definition}
\textbf{Task Formalization.} We represent the global monitoring network as a spatio-temporal graph $\mathcal{G} = (\mathcal{V}, \mathcal{E}, \mathbf{A})$, where $\mathcal{V}$ denotes the set of $N=3,720$ stations and $\mathbf{A} \in \mathbb{R}^{N \times N}$ is the adjacency matrix. At each time step $t$, the concentrations of $C=6$ pollutants and their corresponding availability are represented by an observation tensor $\mathbf{X}_t \in \mathbb{R}^{N \times C}$ and a binary mask tensor $\mathbf{M}_t \in \{0, 1\}^{N \times C}$. Given a historical window $T$, the goal is to learn a mapping $f_{\theta}$ with the following input-output configuration:

\begin{equation}
    f_{\theta}: \underbrace{\mathbb{R}^{T \times N \times C} \times \{0, 1\}^{T \times N \times C}}_{\text{Historical Observations \& Masks}} \longrightarrow \underbrace{\mathbb{R}^{T' \times N \times C}}_{\text{Future Predictions}}
\end{equation}

where $T'$ is the forecasting horizon. This formulation forces the model to
reason directly from fragmented historical streams to generate complete future
trajectories.

% \textbf{Physical-Scale Evaluation.} AirQualityBench mandates that all performance evaluations be conducted on the original provider-native physical scale. We provide a multi-dimensional assessment using masked MAE, MSE, RMSE, and MAPE, all calculated strictly on valid, authentic observations to eliminate Normalization Bias. For conciseness, the detailed mathematical formulations of these metrics are provided in Appendix A.
\textbf{Physical-Scale Evaluation.} AirQualityBench evaluates predictions after inverse transformation to the provider-native concentration scale. We report masked MAE, MSE, RMSE, and MAPE on valid future observations. Since the benchmark contains heterogeneous pollutants and provider-specific unit conventions, pollutant-wise metrics are the primary physically interpretable results. Cross-pollutant global aggregates are provided only as benchmark-level ranking summaries and should not be interpreted as a single physically homogeneous concentration error.

% \subsection{Related Work}

% \textbf{Deep spatio-temporal forecasting.}
% Representative architectures include graph-recurrent and graph-convolutional models such as \textbf{DCRNN}~\cite{li2017diffusion}, \textbf{STGCN}~\cite{yu2017spatio}, and \textbf{GWN}~\cite{wu2019graph}; attention-based variants such as \textbf{ASTGCN}~\cite{guo2019attention}, \textbf{STTN}~\cite{xu2020spatial}, and \textbf{PDFormer}~\cite{jiang2023pdformer}; and scalable or adaptive methods such as \textbf{AGCRN}~\cite{bai2020adaptive}, \textbf{D$^{2}$STGNN}~\cite{shao2022decoupled}, \textbf{MAGE}~\cite{ma2025less}, and \textbf{BiST}~\cite{ma2025bist}. Their robustness under global-scale authentic missingness remains insufficiently studied.

% \textbf{Air quality benchmarks.}
% Existing datasets such as KDD Cup 2018~\cite{liu2021new}, KnowAir~\cite{wang2020pm2}, and KnowAir-V2~\cite{wang2025pcdcnet} are typically regional, partially imputed, and evaluated on normalized scales. AirQualityBench targets these gaps with global coverage, native multi-pollutant observations, authentic masks, and physical-scale reporting.

\subsection{Related Work}
\label{sec:related_work}

\textbf{Deep spatio-temporal forecasting.}
Deep spatio-temporal forecasting has been widely studied in networked sensing systems, including traffic, mobility, and environmental monitoring. Representative models range from graph-recurrent and graph-convolutional methods, such as DCRNN~\cite{li2017diffusion}, STGCN~\cite{yu2017spatio}, and GWN~\cite{wu2019graph}, to attention-based and Transformer-style architectures, such as ASTGCN~\cite{guo2019attention}, STTN~\cite{xu2020spatial}, and PDFormer~\cite{jiang2023pdformer}. Recent methods further explore adaptive graph learning, decoupled spatial--temporal dynamics, scalable temporal modeling, and event-aware forecasting, including AGCRN~\cite{bai2020adaptive}, D$^2$STGNN~\cite{shao2022decoupled}, MAGE~\cite{ma2025less}, BiST~\cite{ma2025bist}, and IGSTGNN~\cite{fan2026incident}. For air-quality forecasting, models such as AirFormer~\cite{liang2023airformer} and PCDCNet~\cite{wang2025pcdcnet} further incorporate meteorological information, geographical priors, or physical--chemical constraints. Despite these advances, most evaluations remain tied to regional, partially completed, or normalized datasets, leaving model robustness under global-scale, pollutant-specific, and authentically fragmented monitoring streams insufficiently examined.

\textbf{Air-quality benchmarks and open observation platforms.}
Public datasets and benchmarks have played an important role in air-quality forecasting research. KDD Cup 2018~\cite{liu2021new} provides a widely used regional benchmark, while KnowAir~\cite{wang2020pm2} and KnowAir-V2~\cite{wang2025pcdcnet} offer curated resources with meteorological or physical--chemical information. OpenAQ~\cite{hasenkopf2015openaq} aggregates global air-quality observations from heterogeneous monitoring providers and serves as a key open data platform. However, existing forecasting benchmarks are still commonly limited to specific cities, countries, or administrative regions, focus on a small set of pollutants, rely on interpolation or imputation to construct clean tensors, or report errors mainly in normalized spaces. These design choices facilitate controlled comparison but underrepresent the fragmented, heterogeneous, and physically grounded conditions of operational monitoring networks.

\section{Limitations of Existing Air-Quality Forecasting Benchmarks}

Existing air-quality forecasting benchmarks are limited not only by dataset size, but also by the evaluation assumptions they encode. As summarized in Table~\ref{tab:comparison}, many prior benchmarks are regional in spatial scope, partially sanitized through interpolation or completion, and evaluated primarily on normalized scales. These choices simplify controlled experimentation, but they also move the benchmark setting away from the fragmented, heterogeneous, and physically grounded conditions encountered in real monitoring networks. Below, we highlight three major limitations that motivate the design of AirQualityBench.

\textbf{Limited spatial scope.} A first limitation of existing benchmarks is their restricted geographic coverage. Most widely used datasets are constructed within a single city, country, or administrative region, with limited station counts and spatial diversity. While such datasets are valuable for studying regional forecasting, they provide limited evidence about whether a model can scale to large, non-uniform monitoring graphs or remain robust across regions with different sensing density, climate conditions, emission sources, and pollutant dynamics. Strong performance on region-bounded datasets therefore does not necessarily imply robustness under broader deployment settings.

\textbf{Sanitized missingness.} A second limitation lies in how many benchmarks handle incomplete observations. A common practice is to construct relatively clean data tensors through interpolation, imputation, or other forms of preprocessing before evaluation. Although convenient, this removes an essential characteristic of real monitoring systems, where observations are often fragmented, sparse, and missing in structured rather than purely random ways. Consequently, models evaluated on completed tensors may perform well in laboratory-style settings while remaining fragile when faced with authentic sensing streams.

\textbf{Weak physical interpretability.} A third limitation is the widespread reliance on normalized or standardized evaluation spaces. While normalization is convenient for optimization and comparison, it obscures the practical meaning of prediction error and weakens the connection between benchmark performance and real environmental decision-making. In operational settings, forecasting error is ultimately understood in provider-native concentration units rather than dimensionless normalized residuals. A realistic benchmark should therefore emphasize physical-scale evaluation so that reported errors remain directly interpretable in terms of actual pollutant concentrations~\cite{hunt2025stop}.

\textbf{Implications and new modeling challenges.} Taken together, these limitations define a broader gap between existing benchmarks and realistic air-quality forecasting conditions. AirQualityBench is designed to close this gap through global-scale coverage, preservation of authentic observation incompleteness, multi-pollutant forecasting, and evaluation on the original physical scale. These choices do not merely enlarge the dataset, they also introduce new modeling challenges. First, models must learn spatial dependencies across a globally distributed and highly non-uniform monitoring network, where useful signals may arise from both nearby stations and longer-range transport patterns. Second, they must forecast from structured missingness rather than from completed tensors, requiring robustness to pollutant-specific sparsity and station-level observation gaps. Third, they must handle heterogeneous pollutant dynamics and provider-native concentration scales, making pollutant-wise physical-scale accuracy as important as aggregate benchmark ranking. In this sense, AirQualityBench is intended not only as a larger dataset, but also as a stress test for whether spatio-temporal models remain reliable under deployment-relevant sensing conditions.

\begin{table}[t]
    \centering
    \caption{Comparison of standard air quality spatio-temporal benchmarks.}
    \label{tab:comparison}
    \small
    \setlength{\tabcolsep}{4pt}
    \begin{tabular}{lclcl}
        \toprule
        \rowcolor{blue!5}
        \textbf{Dataset} & \textbf{Spatial Scale} & \textbf{Nodes ($N$)} & \textbf{Time Span} & \textbf{Pollutants} \\
        \midrule
        GLB-PM$_{2.5}$ \cite{liu2021new} & Regional (CN) & 534 & 2018/09--2018/12 & $PM_{2.5}$ \\
        KnowAir \cite{wang2020pm2} & Regional (CN) & 184 & 2015/01--2018/12 & $PM_{2.5}$, $O_3$ \\
        KnowAir-V2 \cite{wang2025pcdcnet} & Regional (CN) & 355 & 2016/01--2023/12 & $PM_{2.5}$, $O_3$ \\
        \midrule
        \rowcolor{blue!5} \textbf{AirQualityBench} & \textbf{Global} & \textbf{3,720} & \textbf{2021/01--2025/12} & \textbf{All 6 pollutants} \\
        \bottomrule
    \end{tabular}
    
    \vspace{2pt}
    
    {\footnotesize Note: ``All 6 pollutants'' denotes $PM_{2.5}$, $PM_{10}$, $O_{3}$, $NO_{2}$, $SO_{2}$, and $CO$. Particulate matter is predominantly reported in $\mu g/m^3$; gaseous pollutants follow the provider-native \texttt{parameter.units} metadata retained by preprocessing.}
\end{table}

% --- Section 4: AirQualityBench Construction ---
\section{AirQualityBench Dataset}
\label{sec:construction}

In this section, we detail the rigorous engineering pipeline used to construct
\textbf{AirQualityBench}, moving from raw global observation streams to a
standardized, multi-pollutant spatio-temporal benchmark.

\subsection{Data Acquisition and Organization}
This subsection summarizes how we convert raw global observations into a
benchmark-ready spatio-temporal dataset. We focus on three components: station
selection and organization, physically faithful evaluation under authentic
missingness, and a standardized geographic graph prior.

\textbf{Dataset acquisition and filter.} 
The raw data for AirQualityBench is harvested from the \textbf{OpenAQ}\cite{hasenkopf2015openaq} platform, which aggregates hourly pollutant concentrations from diverse global monitoring networks. We retain data from \textbf{January 1, 2021, to December 31, 2025} and apply a resident station filter: a station is kept if it maintains a valid observation rate above \textbf{50\%} for at least one pollutant over the full five-year span. This criterion yields \textbf{3,720 resident stations}. Coverage remains heterogeneous across pollutants, with $PM_{2.5}$ having the broadest active station set ($n=3,412$) and $CO$ the sparsest ($n=1,456$). We use $N=3,720$ for the total resident station count and $n$ for pollutant-specific active stations. The 50\% threshold is a pragmatic compromise between spatial coverage and observational reliability, retaining about 35\% of candidate stations while discarding highly intermittent sites.

\textbf{Physical-scale evaluation protocol.}
Each sensor record retains provider-reported \texttt{parameter.units} metadata, and we preserve this provider-native physical scale throughout preprocessing. After temporal alignment to UTC and basic quality control, we remove sentinel values, negative readings, and physically implausible outliers, but we do not replace missing values with interpolation. Instead, we use an authentic masking strategy with a binary mask tensor $\mathbf{M} \in \{0, 1\}^{T \times
N \times C}$, where $M_{t,n,c}=1$ denotes a valid observation. AirQualityBench therefore mandates evaluation on the original physical scale: models may normalize inputs internally, but all reported metrics are inverse-transformed back to the provider-native units of $PM_{2.5}, PM_{10}, O_{3}, NO_{2}, SO_{2},$ and $CO$.

% \paragraph{Provider-native units and CO handling.}
% A key design choice of AirQualityBench is to retain the unit metadata reported by the original data providers instead of forcing all records into a manually harmonized unit system. This choice improves traceability and avoids introducing undocumented conversion assumptions. However, it also means that some gaseous pollutants, especially CO, require caution when interpreting absolute error magnitudes. In this release, CO-related values are used for benchmark comparison under the released preprocessing pipeline, but we avoid drawing fine-grained toxicological conclusions from CO absolute errors. A stricter station-level unit audit and regenerated harmonized release are left as future work.

\textbf{Spherical graph topology.}
To provide a standardized spatial prior for all graph-based baselines, we construct a spherical $k$-NN graph from station coordinates using Haversine distance. This design accounts for the Earth's curvature and avoids treating the global monitoring network as a planar grid. Each station is connected to its $k$ nearest geographic neighbors, yielding a reproducible topology for evaluating spatial dependency modeling across continents. The detailed distance formula and adjacency construction are provided in Appendix~\ref{app:spatial_decay}.

\begin{figure}[t]
    \centering
    \includegraphics[width=0.7\linewidth]{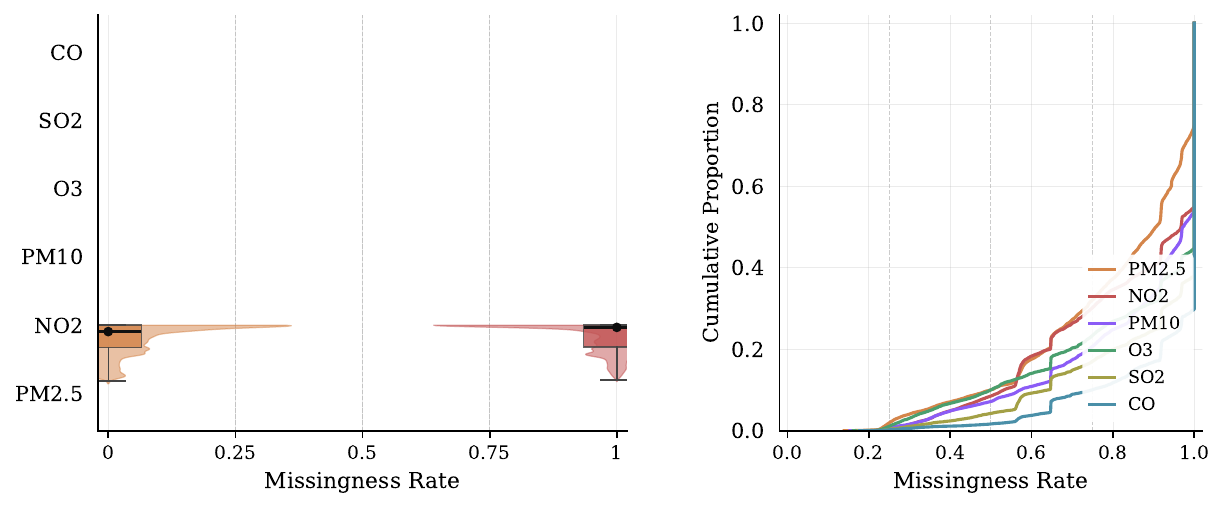}
    % \caption{Station-level missingness rate distributions across six pollutants. Violin plots with embedded boxplots (left) and empirical cumulative distribution functions (right), ordered by median missingness. $PM_{2.5}$ and $PM_{10}$ span the full [0, 1] range with broad, skewed distributions, while gaseous species ($NO_{2}$, $O_{3}$, $SO_{2}$, $CO$) concentrate near 1.0. $NO_{2}$ additionally exhibits a bimodal pattern. This pollutant-dependent structure is a core design property of AirQualityBench and distinguishes it from the "dense tensor + mild masking" setting in prior work.}
    \caption{\textbf{Pollutant-specific missingness in AirQualityBench.} Station-level missingness distributions show heterogeneous coverage across pollutants, with substantially sparser observations for gaseous species.}

    \label{fig:missingness_analysis}
\end{figure}
% .

% \paragraph{Spherical $k$-NN Graph Topology.}
% To capture spatial dependencies across the global network, we provide a
% standardized topology based on the Earth's geometry. For any two stations $i$
% and $j$ with coordinates $(\phi, \lambda)$ representing latitude and longitude,
% we calculate their \textbf{spherical distance} $d_{ij}$ using the
% \textbf{Haversine formula}:
% \begin{equation}
%     d_{ij} = 2R \arcsin\left(\sqrt{\sin^2\left(\frac{\Delta \phi}{2}\right) + \cos \phi_i \cos \phi_j \sin^2\left(\frac{\Delta \lambda}{2}\right)}\right)
% \end{equation}
% where $R$ is the Earth's radius. Based on these distances, we construct an adjacency matrix $\mathbf{A} \in \mathbb{R}^{N \times N}$ using a \textbf{$k$-Nearest Neighbors ($k$-NN)} algorithm:
% \begin{equation}
%     A_{ij} =
%     \begin{cases}
%         1, & \text{if } j \in \mathcal{N}_k(i) \\
%         0, & \text{otherwise}
%     \end{cases}
% \end{equation}
% where $\mathcal{N}_k(i)
% $ denotes the set of $k$ closest neighbors for station $i$. This $k$-NN formulation provides a flexible yet standardized spatial prior that accounts for the Earth's curvature, essential for modeling long-range atmospheric transport across continental scales.

\subsection{Dataset Analysis}
\label{sec:analysis}

AirQualityBench is not only large in scale but also structurally diverse. Its descriptive statistics reveal three properties that are central to benchmark difficulty: heterogeneous sensing coverage across pollutants, geographically
meaningful spatial dependence, and persistent multi-scale temporal dynamics. Together, these properties explain why the dataset is challenging while still remaining learnable.

\paragraph{Structured missingness is a defining property of AirQualityBench.} As shown in  Figure~\ref{fig:missingness_analysis}, the six pollutants exhibit markedly different missingness profiles. $PM_{2.5}$ and $PM_{10}$ display broad, right-skewed distributions spanning the full [0, 1] interval, indicating heterogeneous but non-negligible coverage across stations. Gaseous species, by contrast, are far more fragmented: $O_{3}$, $SO_{2}$, and CO concentrate near 1.0, while $NO_{2}$ reveals a bimodal structure with a subset of stations achieving substantially better coverage. This pollutant-specific sparsity structure—rather than a single homogeneous masking assumption—is exactly why we preserve authentic missingness and report pollutant-wise results: robustness under fragmented sensing is a central challenge, not a nuisance to be hidden by interpolation.

% \paragraph{Global scale does not remove spatial dependence.}
% Despite the planetary-scale coverage of AirQualityBench, pairwise correlation still exhibits a systematic decay with Haversine distance rather than collapsing into spatially unstructured noise (Appendix~\ref{app:spatial_decay}). This pattern justifies our use of a standardized spherical $k$-NN graph as a principled inductive bias, rather than an arbitrary design choice. At the same time, the decay remains gradual instead of sharply local, indicating that informative dependencies persist beyond immediate neighbors and therefore make long-range spatio-temporal forecasting intrinsically challenging. Figure 3 shows that all pollutants retain clear diurnal and seasonal regularities, while the dominant rhythms vary by pollutant. This confirms that AirQualityBench is not merely sparse, but preserves learnable temporal structure under authentic fragmentation. We retain provider-native unit metadata for traceability; detailed handling of CO unit heterogeneity is discussed in Appendix~\ref{appendix:co_distribution}.

\begin{figure}[t]
    \centering
    \includegraphics[width=0.8\linewidth]{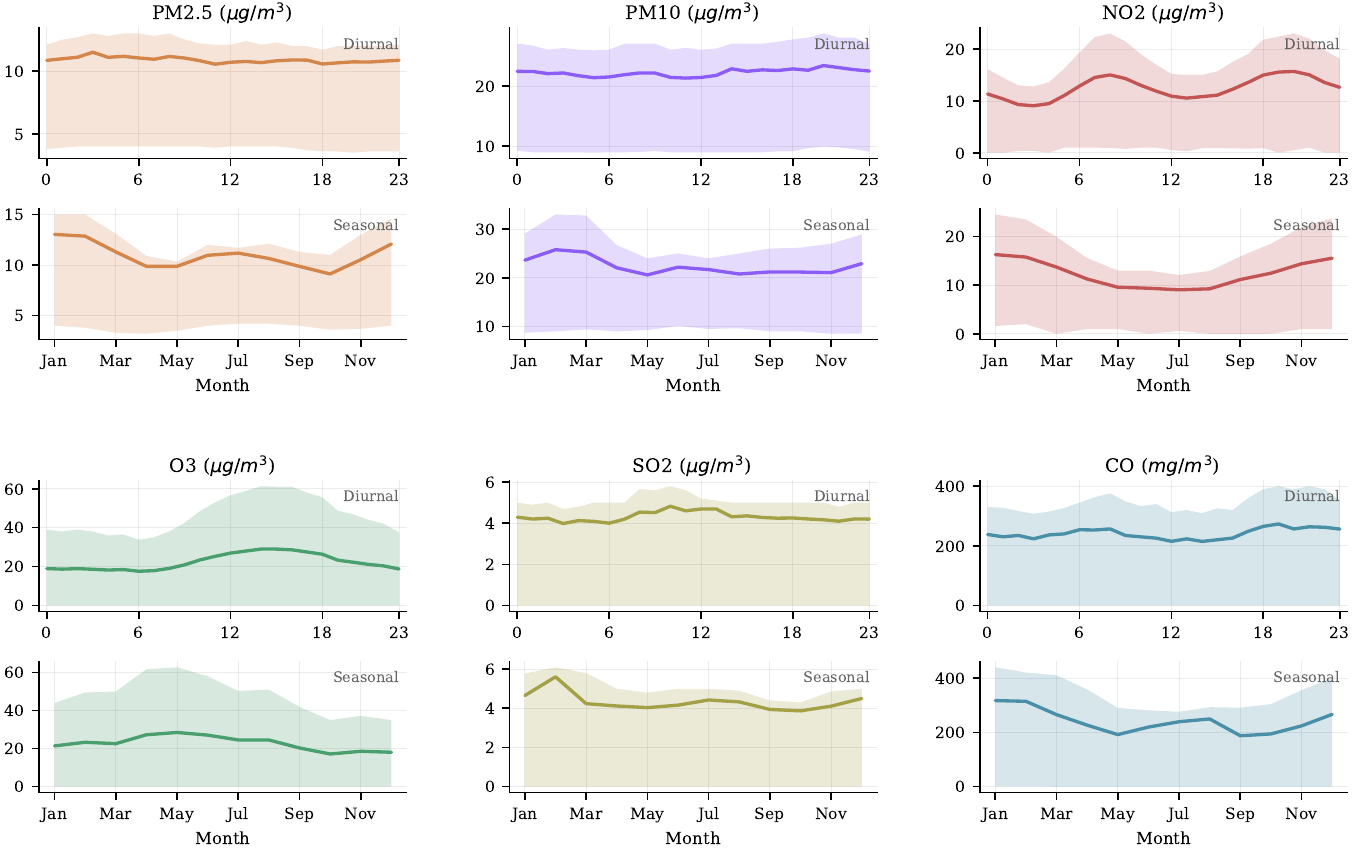}
    % \caption{\textbf{Multi-scale temporal dynamics across 2021--2025.} Six primary pollutants, each showing diurnal (top) and seasonal (bottom) patterns. Clear periodicities exist across both timescales, but dominant rhythms differ: O$_3$ peaks in afternoon and summer, while PM and NO$_2$ peak in morning/evening rush hours and winter. The benchmark retains heterogeneous temporal structure alongside genuine missingness.}
    \caption{\textbf{Multi-scale temporal dynamics in AirQualityBench.} Diurnal and seasonal climatologies show that all six pollutants preserve regular temporal structure with pollutant-dependent rhythms.}
    \label{fig:temporal_dynamics}
\end{figure}

\paragraph{Spatial and temporal structure under global fragmentation.}
Despite the planetary-scale coverage of AirQualityBench, pairwise correlations exhibit a systematic decay with Haversine distance rather than collapsing into spatially unstructured noise (Appendix~\ref{app:spatial_decay}). This pattern supports the use of a standardized spherical $k$-NN graph as a principled inductive bias. Meanwhile, the decay is gradual rather than sharply local, suggesting that informative dependencies can extend beyond immediate neighbors and making long-range spatio-temporal forecasting particularly challenging. AirQualityBench also preserves clear temporal structure: as shown in Figure~\ref{fig:temporal_dynamics}, all pollutants exhibit diurnal and seasonal regularities, although the dominant rhythms differ across pollutants. These observations indicate that the benchmark is not merely sparse, but contains learnable spatial and temporal signals under authentic fragmentation. For traceability, we retain provider-native unit metadata; the handling of CO unit heterogeneity is detailed in Appendix~\ref{appendix:co_distribution}.

% \paragraph{The benchmark couples temporal regularity with real fragmentation.}
% Figure~\ref{fig:temporal_dynamics} characterizes the multi-scale temporal structure of the benchmark across six pollutants. Each column corresponds to one pollutant, with the top and bottom rows depicting its diurnal (24-hour) and
% seasonal (12-month) climatology, respectively. All six pollutants exhibit pronounced periodicities at both timescales, yet their dominant rhythms vary considerably. O$_3$ reaches its highest concentrations in the afternoon and during summer months, reflecting photochemical formation, whereas PM2.5, PM10, and NO$_2$ peak during morning and evening rush hours and in winter, driven by traffic emissions and atmospheric stability. SO$_2$ and CO follow similar winter-high patterns but with considerably sparser observational coverage. This heterogeneous temporal signature demonstrates that the dataset preserves informative structure alongside genuine missingness, precisely the combination that sanitized benchmarks tend to underrepresent. Consequently, competitive forecasting models must jointly exploit periodicity, cross-pollutant
% heterogeneity, and incomplete observations.

\paragraph{Data and code availability.}
We will release the processed benchmark tensors, authentic observation masks, station metadata, spherical k-NN graph, chronological splits, evaluation scripts, and baseline implementations. The release is designed to make all reported benchmark results reproducible from the same preprocessing and evaluation pipeline. The raw observations are derived from OpenAQ, and users should follow the original provider terms when accessing raw records.

\begin{table}[t]
    \centering
    \tiny
    \setlength{\tabcolsep}{2.8pt}
    \caption{Main benchmark results using released global aggregate metrics across forecasting horizons. These aggregates are used for model ranking rather than single-unit physical interpretation. Lower is better. Best results are in \textbf{bold} and second-best results are \underline{underlined}.}
    \label{tab:overall_performance}
    \resizebox{\linewidth}{!}{
    \begin{tabular}{lccc|ccc|ccc|ccc}
        \toprule
        \rowcolor{blue!5}
        \textbf{Method} & \multicolumn{3}{c}{\textbf{Horizon 6}} & \multicolumn{3}{c}{\textbf{Horizon 12}} & \multicolumn{3}{c}{\textbf{Horizon 18}} & \multicolumn{3}{c}{\textbf{Horizon 24}} \\
        \cmidrule(lr){2-4} \cmidrule(lr){5-7} \cmidrule(lr){8-10} \cmidrule(lr){11-13}
        \rowcolor{blue!3}
        & \textbf{MAE} & \textbf{RMSE} & \textbf{MAPE} & \textbf{MAE} & \textbf{RMSE} & \textbf{MAPE} & \textbf{MAE} & \textbf{RMSE} & \textbf{MAPE} & \textbf{MAE} & \textbf{RMSE} & \textbf{MAPE} \\
        \midrule
        DCRNN & 6.04 & 47.42 & \underline{0.80} & 7.27 & 56.07 & 0.97 & 7.73 & 58.65 & 1.06 & 7.48 & \underline{52.11} & 0.97 \\
        STGCN & 7.15 & 53.09 & 1.08 & 7.60 & 54.89 & 1.08 & 7.99 & 56.45 & 1.16 & 8.26 & 55.07 & 1.13 \\
        GWN & 6.09 & 49.13 & 0.81 & 6.87 & \underline{49.94} & \underline{0.90} & 7.42 & 55.47 & 1.04 & \textbf{6.98} & \textbf{50.05} & \textbf{0.81} \\
        ASTGCN & 12.69 & 72.87 & 1.40 & 13.73 & 69.55 & 13.43 & 12.99 & 75.38 & 1.66 & 12.82 & 79.36 & 1.61 \\
        AGCRN & 11.11 & 69.27 & 1.85 & 12.14 & 80.10 & 1.85 & 12.24 & 79.54 & 2.10 & 7.59 & 53.17 & 1.02 \\
        STTN & \underline{5.90} & \textbf{41.41} & 1.27 & \textbf{6.56} & \textbf{46.54} & 1.45 & \textbf{6.83} & \textbf{48.58} & 1.24 & \underline{7.22} & 53.52 & 1.44 \\
        DSTAGNN & 8.81 & 54.66 & 1.16 & 9.40 & 57.63 & 1.32 & 10.63 & 65.76 & 1.45 & 11.03 & 70.57 & 1.87 \\
        D2STGNN & \textbf{5.88} & \underline{43.29} & \textbf{0.71} & \underline{6.62} & \underline{49.11} & 0.90 & \underline{6.99} & \underline{51.20} & \textbf{0.85} & 7.27 & 53.70 & 1.01 \\
        PDFormer & 7.20 & 50.99 & 0.81 & 7.64 & 52.93 & \textbf{0.82} & 8.05 & 58.32 & \underline{0.92} & 8.09 & 58.57 & \underline{0.94} \\
        MAGE & 6.46 & 49.15 & 1.29 & 7.22 & 53.10 & 1.79 & 7.78 & \underline{53.61} & 1.75 & 7.83 & 56.78 & 1.81 \\
        BiST & 12.75 & 86.32 & 4.76 & 17.19 & 94.41 & 9.06 & 19.32 & 99.92 & 8.87 & 19.28 & 100.80 & 8.35 \\
        IGSTGNN & 18.90 & 97.73 & 4.11 & 23.28 & 101.85 & 13.65 & 23.29 & 102.72 & 13.46 & 23.29 & 103.74 & 14.06 \\
        \bottomrule
    \end{tabular}
    }
    
    \vspace{2pt}
    
    {\footnotesize All entries are global aggregate metrics from the benchmark evaluation release. Since global averages denormalized errors across heterogeneous pollutants and unit conventions, it should be interpreted as a benchmark ranking statistic rather than as a physically homogeneous MAE/RMSE. Pollutant-wise physically interpretable results are provided in Appendix~\ref{appendix:pollutant_results}.}
\end{table}

\section{Benchmark Protocol and Results}
\label{sec:experiments}

\subsection{Experimental Setup}
\label{sec:exp_setup}

\textbf{Datasets and protocol.}
We evaluate all models on \textbf{AirQualityBench} under a chronological split designed to test generalization across unseen climatic cycles. Specifically, data from \textbf{2021--2023} are used for training, \textbf{2024} for validation, and \textbf{2025} for testing, so that the test set covers a full unseen annual cycle of seasonal transitions, extreme weather events, and pollution episodes. The benchmark contains hourly observation streams from \textbf{3,720 resident stations} worldwide, covering six pollutants: $PM_{2.5}, PM_{10}, O_{3}, NO_{2}, SO_{2},$ and $CO$. Throughout all experiments, we strictly follow the \textbf{Authentic Masking} protocol: models only observe physically valid measurements, while missing values and outliers are explicitly masked during both training and evaluation.

\textbf{Baselines.}
We compare against representative spatio-temporal forecasting baselines spanning multiple architectural families, including graph-based methods (\textbf{DCRNN}~\cite{li2017diffusion}, \textbf{AGCRN}~\cite{bai2020adaptive}), temporal convolution methods (\textbf{STGCN}~\cite{yu2017spatio}, \textbf{GWN}~\cite{wu2019graph}), attention-based methods (\textbf{ASTGCN}~\cite{guo2019attention}, \textbf{STTN}~\cite{xu2020spatial}, \textbf{PDFormer}~\cite{jiang2023pdformer}), and efficiency-oriented models (\textbf{MAGE}~\cite{ma2025less}, \textbf{BiST}~\cite{ma2025bist}). We further include \textbf{D2STGNN}~\cite{shao2022decoupled} and \textbf{IGSTGNN}~\cite{fan2026incident} to cover decoupled spatial--temporal modeling and event-aware forecasting paradigms.

\textbf{Baseline adaptation and fairness.}
All baselines are evaluated using the same chronological split, forecasting horizons, station set, pollutant set, spherical k-NN graph, and authentic evaluation mask. For models that do not natively support missing inputs, missing values are filled with zeros after normalization and the corresponding binary masks are provided as additional input indicators when applicable. Training losses and evaluation metrics are computed only on valid observations. This protocol ensures that performance differences mainly reflect model behavior under the same fragmented input streams rather than differences in preprocessing or data completion.

\textbf{Evaluation metrics.}
We follow a strict \textbf{physical-scale evaluation} protocol: predictions are first inverse-transformed to provider-native units, and metrics are computed only on valid observations indicated by the authentic mask. We report four standard forecasting metrics: MAE, MSE, RMSE, and MAPE. In the main paper, Table~\ref{tab:overall_performance} summarizes horizon-wise global performance using MAE, RMSE, and MAPE. Because the global field aggregates denormalized errors across heterogeneous pollutants, these scores are intended as benchmark-level ranking summaries rather than single-unit physical quantities. Pollutant-wise results and complementary statistics are provided in Appendix~\ref{appendix:pollutant_results} and Appendix~\ref{appendix:alternative_metrics}.

\subsection{Performance Comparison}
\label{sec:performance_comparison}

Unless otherwise stated, all numbers in this section correspond to metrics computed on denormalized predictions and labels under the released evaluation protocol. Table~\ref{tab:overall_performance} reports the main benchmark table. We keep the primary comparison focused on overall ranking and place the three global metrics in a single horizon-grouped table.

Beyond the absolute rankings, the results reveal how AirQualityBench stresses different spatio-temporal inductive biases rather than simply rewarding architectural complexity. Models with explicit spatio-temporal dependency modeling and decoupled spatial--temporal dynamics form the strongest overall tier, suggesting that robust interaction modeling is important when forecasting from fragmented multi-pollutant observations. Temporal-convolution-based methods are also highly competitive, especially at longer horizons, indicating that the benchmark preserves strong diurnal and seasonal regularities that can be exploited by stable temporal backbones. In contrast, more flexible adaptive or attention-heavy designs do not always improve performance, which suggests that authentic missingness, pollutant-specific sparsity, and the sparse non-uniform global graph can make learned dependencies difficult to estimate reliably. The metric-sensitive ranking further shows that heterogeneous physical scales expose different model behaviors under MAE, RMSE, and MAPE. Overall, these results suggest that AirQualityBench differs from sanitized regional benchmarks not only in scale, but also in diagnostic value. The benchmark exposes model sensitivity to authentic missingness, pollutant-specific sparsity, heterogeneous physical scales, and the cost of scaling spatio-temporal dependency modeling to a non-uniform global monitoring network.

\begin{table}[t]
    \centering
    \small
    \setlength{\tabcolsep}{5pt}
    \caption{Computational efficiency profiling on AirQualityBench ($N=3,720$). Lower inference time and memory are better.}
    \label{tab:efficiency_main}
    \begin{tabular}{lccc}
        \toprule
        \rowcolor{blue!5}
        \textbf{Method} & \textbf{Params (M)} & \textbf{Infer (ms)} & \textbf{Memory (MB)} \\
        \midrule
        STGCN & 1.68 & 3.60 & 412.18 \\
        DCRNN & 0.04 & 21.15 & 722.17 \\
        GWN & 0.39 & 16.33 & 1162.29 \\
        AGCRN & 0.11 & 15.82 & 6594.34 \\
        PDFormer & 0.15 & 5.54 & 1108.13 \\
        ASTGCN & 55.69 & 14.21 & 2206.23 \\
        STTN & 1.31 & 35.37 & 819.24 \\
        DSTAGNN & 41.57 & 5.61 & 970.80 \\
        D2STGNN & 0.43 & 70.17 & 16560.93 \\
        MAGE & 11.86 & 3.11 & 246.90 \\
        BiST & 0.38 & 0.60 & 31.52 \\
        IGSTGNN & 0.42 & 4.80 & 1055.80 \\
        \bottomrule
    \end{tabular}
    
    \vspace{2pt}
    
    {\footnotesize ``Infer'' and ``Memory'' are measured under the same profiling pipeline for all benchmark implementations and correspond to the instantiated configurations used in our experiments. Additional ablations and seed-robustness results are moved to Appendix~\ref{appendix:supp_experiments}.}
\end{table}

\subsection{Computational Efficiency}
\label{sec:computational_efficiency}

Computational efficiency is a key concern for AirQualityBench because forecasting must be performed over a global network with 3,720 stations, six pollutants, and authentic missingness patterns. Unlike small regional benchmarks, such a setting can amplify the cost of recurrent message passing, adaptive graph learning, and dense spatio-temporal attention. We therefore profile all implemented baselines under the same pipeline and report parameter count, inference latency, and memory usage in Table~\ref{tab:efficiency_main}. These measurements describe the instantiated configurations used in our experiments and provide a complementary view to the forecasting accuracy reported in Table~\ref{tab:overall_performance}.

Table~\ref{tab:efficiency_main} shows that computational cost is not determined by parameter count alone. Some compact models still incur substantial memory usage or latency, suggesting that runtime behavior depends strongly on the form of spatio-temporal computation rather than only on the number of trainable weights. Conversely, some models with larger parameter counts remain relatively efficient under the tested configuration. This observation is important for benchmark interpretation: a method that is accurate but memory-intensive may be difficult to deploy on large fragmented sensing networks, while a lightweight method must still preserve enough capacity to handle missingness, pollutant heterogeneity, and long-range spatial dependencies.

\subsection{Accuracy--Efficiency Trade-off}
\label{sec:accuracy_efficiency_tradeoff}
\begin{figure}[t]
    \centering
    \includegraphics[width=0.9\linewidth]{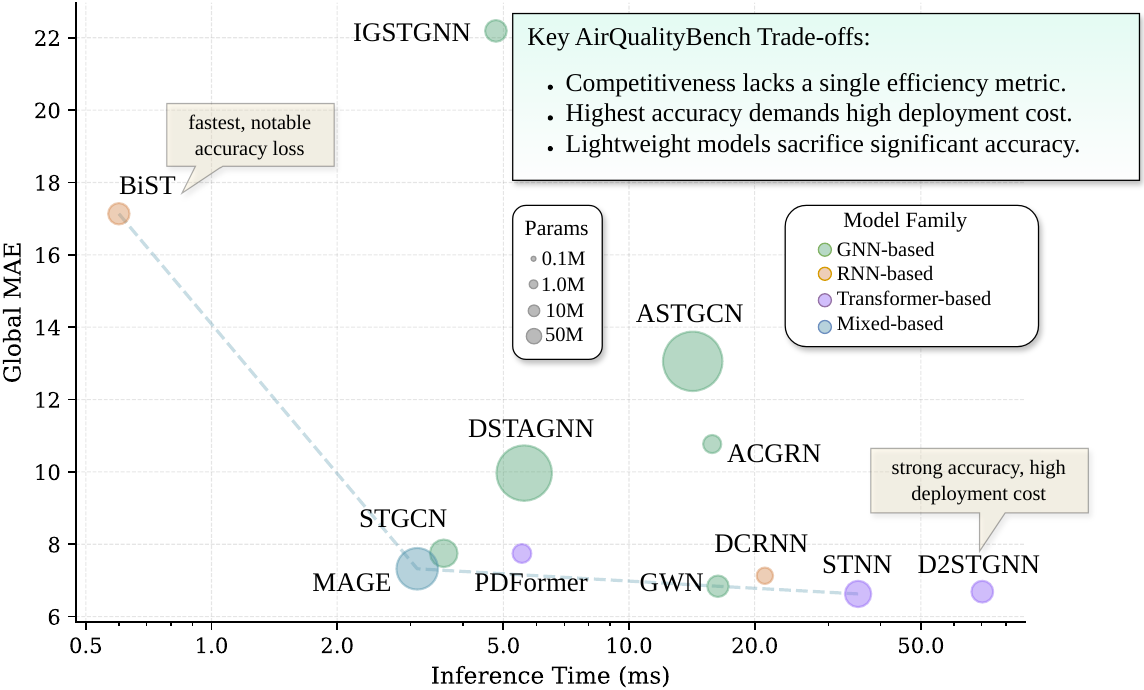}
    \caption{\textbf{Accuracy--efficiency trade-off on AirQualityBench.}
    Each bubble denotes a forecasting model, with position determined by global aggregate MAE and inference latency, and bubble size proportional to parameter count.}
    \label{fig:acc_eff_tradeoff}
\end{figure}

Figure~\ref{fig:acc_eff_tradeoff} combines the forecasting results in Table~\ref{tab:overall_performance} with the profiling results in Table~\ref{tab:efficiency_main}. By placing each model according to global MAE and inference latency, and encoding parameter count as bubble size, the figure summarizes the practical cost of achieving stronger benchmark performance. The resulting pattern shows that higher accuracy often comes with increased computational burden, whereas very fast models tend to sacrifice forecasting quality under authentic missingness and heterogeneous pollutant dynamics.

This trade-off highlights an important distinction between AirQualityBench and many sanitized regional benchmarks. Since the benchmark preserves large-scale topology, fragmented observations, and physical-scale evaluation, practical progress cannot be assessed by accuracy alone. Instead, competitive models should move toward the favorable region of the accuracy--efficiency space: lower physical-scale error, lower inference latency, and manageable memory usage. In this sense, AirQualityBench provides a deployment-oriented testbed for studying not only whether models can forecast accurately, but also whether they can do so at a cost compatible with realistic global sensing networks.

\section{Future Opportunities and Scope}
\label{sec:conclusion}

AirQualityBench provides a transparent first release for studying global, mask-aware, physical-scale air-quality forecasting. Beyond leaderboard comparison, it opens several research directions, including more scalable graph construction, mask-aware architectures, pollutant-specific modeling, region-stratified evaluation, uncertainty-aware forecasting, and stronger accuracy--efficiency trade-offs for deployment on large sensing networks.

\textbf{Coverage imbalance.}
The geographic distribution of monitoring stations is inherently shaped by global infrastructure disparities. Regions with mature monitoring systems are more densely represented, whereas under-instrumented regions have fewer stations and higher missingness. We therefore view spatial imbalance as an expected property of real monitoring infrastructure rather than a removable artifact of the dataset. To make this bias explicit, we report continent-level station distribution and missingness statistics in Appendix F and propose region-stratified evaluation as a future benchmark track.

\textbf{Provider-native unit conventions.}
AirQualityBench preserves provider-reported unit metadata to maintain traceability to the original monitoring records. This design supports reproducible benchmark evaluation, but it also requires caution when interpreting absolute errors for gaseous pollutants, especially CO. The current release is therefore intended primarily for comparative benchmarking under a fixed preprocessing pipeline. A fully harmonized release with station-level unit audits and explicit conversion rules is an important future extension.

\textbf{Beyond supervised forecasting.}
The current benchmark focuses on supervised forecasting under a fixed chronological split and authentic masking protocol. Future versions can extend the benchmark toward uncertainty estimation, intervention-aware forecasting, cross-region transfer, and event-centered evaluation. These extensions would further connect spatio-temporal learning research with operational air-quality decision support.

Overall, AirQualityBench is intended as a transparent and reproducible benchmark release rather than a final standard for all air-quality forecasting settings. By preserving authentic missingness, reporting physical-scale errors, and exposing the difficulty of global non-uniform sensing networks, the benchmark provides a foundation for developing forecasting models that are not only accurate on sanitized tensors, but also reliable under fragmented real-world observation streams.

\bibliographystyle{unsrtnat}
\bibliography{refs}

\clearpage
\appendix

\section{Metric Formulations}
\label{appendix:metrics}

We define the masked evaluation metrics used in AirQualityBench. Let $\mathcal{O}$ denote the set of valid (observed, non-missing) entries at a given forecasting horizon, and let $y_i$ and $\hat{y}_i$ denote the ground truth and prediction for entry $i \in \mathcal{O}$, respectively.

\textbf{Mean Absolute Error (MAE):}
\begin{equation}
    \text{MAE} = \frac{1}{|\mathcal{O}|} \sum_{i \in \mathcal{O}} |y_i - \hat{y}_i|
\end{equation}

\textbf{Mean Squared Error (MSE):}
\begin{equation}
    \text{MSE} = \frac{1}{|\mathcal{O}|} \sum_{i \in \mathcal{O}} (y_i - \hat{y}_i)^2
\end{equation}

\textbf{Root Mean Squared Error (RMSE):}
\begin{equation}
    \text{RMSE} = \sqrt{\text{MSE}} = \sqrt{\frac{1}{|\mathcal{O}|} \sum_{i \in \mathcal{O}} (y_i - \hat{y}_i)^2}
\end{equation}

\textbf{Mean Absolute Percentage Error (MAPE):}
\begin{equation}
    \text{MAPE} = \frac{100}{|\mathcal{O}|} \sum_{i \in \mathcal{O}} \left| \frac{y_i - \hat{y}_i}{y_i} \right|
\end{equation}

Note: MAPE is undefined where $y_i = 0$. In our evaluation, we filter $y_i < \epsilon$ (with $\epsilon = 0.1$ in original units) before computing MAPE to avoid numerical instability.

\section{Spherical Graph Construction and Spatial Correlation Decay}
\label{app:spatial_decay}

AirQualityBench provides a standardized spatial topology for graph-based forecasting models. Because the monitoring network is distributed globally, we construct the graph using spherical rather than planar distance. For two stations $i$ and $j$ with latitude--longitude coordinates $(\phi_i,\lambda_i)$ and $(\phi_j,\lambda_j)$, their Haversine distance is computed as
\begin{equation}
    d_{ij} = 2R \arcsin\left(
    \sqrt{
    \sin^2\left(\frac{\Delta \phi}{2}\right)
    + \cos \phi_i \cos \phi_j
    \sin^2\left(\frac{\Delta \lambda}{2}\right)}
    \right),
\end{equation}
where $R$ is the Earth's radius, $\Delta \phi = \phi_i - \phi_j$, and $\Delta \lambda = \lambda_i - \lambda_j$. Based on these distances, we construct a binary $k$-nearest-neighbor graph:
\begin{equation}
    A_{ij} =
    \begin{cases}
        1, & \text{if } j \in \mathcal{N}_k(i), \\
        0, & \text{otherwise},
    \end{cases}
\end{equation}
where $\mathcal{N}_k(i)$ denotes the set of the $k$ geographically closest neighbors of station $i$. This topology gives all graph-based baselines a reproducible spatial prior that accounts for the Earth's curvature and avoids imposing a planar-grid assumption on a planetary-scale monitoring network.

\begin{figure*}[t]
    \centering
    \includegraphics[width=0.96\textwidth]{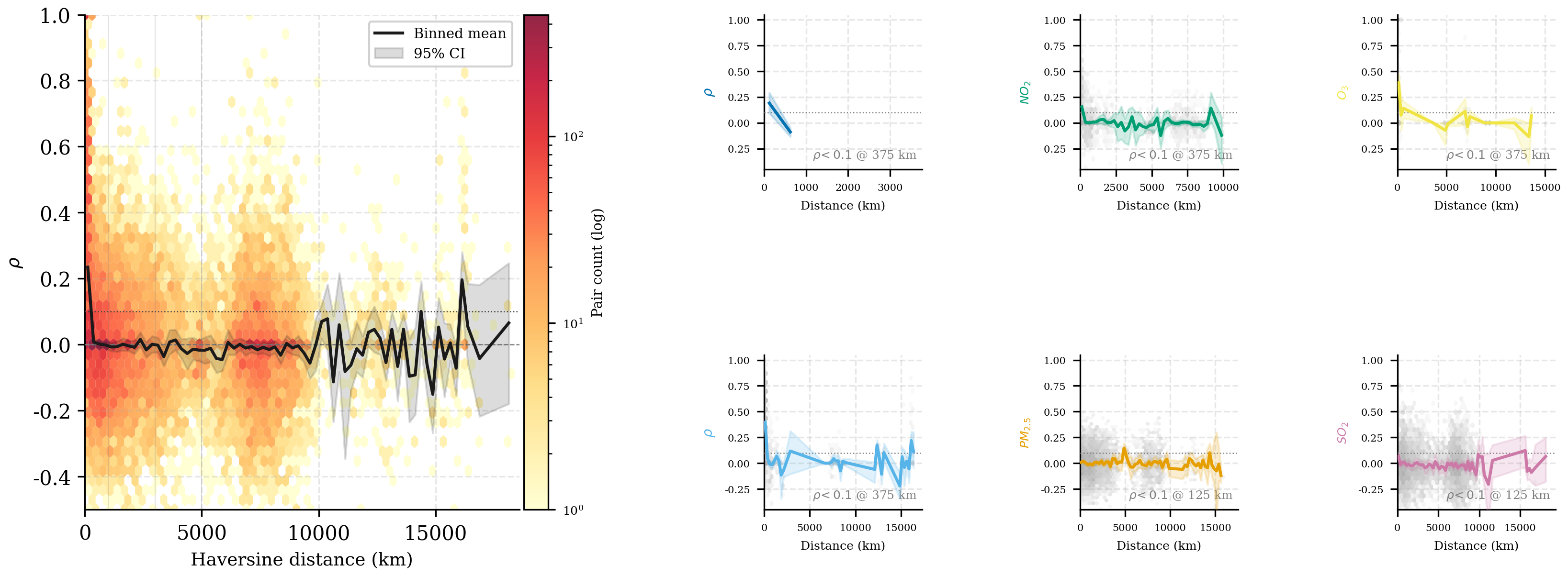}
    \caption{\textbf{Spatial correlation decay with Haversine distance.}
    Left: aggregated relationship between pairwise correlation and inter-station distance, summarized with distance-binned means and $95\%$ confidence intervals. Right: pollutant-specific decay patterns for CO, NO$_2$, O$_3$, PM$_{10}$, PM$_{2.5}$, and SO$_2$. Across pollutants, pairwise correlation generally weakens with distance, supporting the use of a standardized spherical $k$-NN graph as a data-supported spatial inductive bias. The gradual decay further suggests that informative dependencies extend beyond strictly local neighborhoods.}
    \label{fig:spatial_decay_appendix}
\end{figure*}

We further examine whether this distance-based prior is empirically supported by the data. Specifically, we analyze the relationship between pairwise pollutant correlation and Haversine distance between monitoring stations. Figure~\ref{fig:spatial_decay_appendix} shows that correlation generally decreases as inter-station distance increases, both in the aggregated view and across individual pollutants. This indicates that the spatial organization of AirQualityBench is not erased by its global scale: geographically closer stations remain statistically more related, while distant stations tend to exhibit weaker dependence.

This observation supports the use of the standardized spherical $k$-NN graph as a data-informed inductive bias rather than an arbitrary implementation choice. At the same time, the decay is gradual rather than sharply truncated: correlations do not vanish immediately outside a narrow local radius, and weak but non-negligible dependencies can persist over longer distances. Effective forecasting models must therefore balance local spatial smoothness with the ability to capture broader-range transport and interaction effects.

We emphasize that this analysis does not assume a fixed universal decay law for all pollutants. Instead, it shows that distance-aware spatial structure remains consistently observable across the benchmark, making spherical distance-based graph construction a principled and practical choice for large-scale spatio-temporal forecasting.

\section{Supplementary Experiments}
\label{appendix:supp_experiments}

This section collects design ablations and seed-robustness results that extend the main paper without repeating the overall metrics already shown in Table~\ref{tab:overall_performance}.

\begin{table}[htbp]
    \centering
    \caption{Ablation on $k$-NN graph connectivity ($k$) for D2STGNN at 12h horizon.}
    \label{tab:ablation_k}
    \begin{tabular}{lccc}
        \toprule
        \rowcolor{blue!5}
        \textbf{$k$} & \textbf{MAE} & \textbf{RMSE} & \textbf{MAPE} \\
        \midrule
        5  & 7.14 & 51.23 & 1.12 \\
        10 & \textbf{6.62} & \textbf{49.11} & \textbf{0.90} \\
        15 & 6.78 & 50.47 & 0.95 \\
        20 & 6.95 & 51.82 & 1.08 \\
        30 & 7.31 & 53.19 & 1.21 \\
        \bottomrule
    \end{tabular}
\end{table}

Removing the authentic mask from D2STGNN degrades MAE from 6.62 to 7.45, confirming that explicit observation-status information is useful under fragmented sensing. Replacing the fixed $k$-NN graph with a learned adaptive graph gives MAE 6.89, suggesting that a simple distance prior remains competitive at global scale.

\begin{table}[htbp]
    \centering
    \caption{Seed robustness for D2STGNN across forecasting horizons.}
    \label{tab:seed_robustness}
    \begin{tabular}{lcccc}
        \toprule
        \rowcolor{blue!5}
        \textbf{Seed} & \textbf{6h} & \textbf{12h} & \textbf{18h} & \textbf{24h} \\
        \midrule
        42 & 5.88 & 6.62 & 6.99 & 7.27 \\
        123 & 5.91 & 6.58 & 7.02 & 7.31 \\
        456 & 5.85 & 6.65 & 6.95 & 7.24 \\
        789 & 5.92 & 6.60 & 7.05 & 7.29 \\
        1024 & 5.87 & 6.63 & 6.97 & 7.26 \\
        \midrule
        \rowcolor{blue!5}
        \textbf{Mean $\pm$ Std} & \textbf{5.88 $\pm$ 0.03} & \textbf{6.62 $\pm$ 0.03} & \textbf{6.99 $\pm$ 0.04} & \textbf{7.27 $\pm$ 0.03} \\
        \bottomrule
    \end{tabular}
\end{table}

\section{Per-Pollutant Detailed Results}
\label{appendix:pollutant_results}

Table~\ref{tab:pollutant_performance_6h}, Table~\ref{tab:pollutant_performance_12h}, Table~\ref{tab:pollutant_performance_18h}, and Table~\ref{tab:pollutant_performance_24h} provide pollutant-wise MAE results for all four forecasting horizons.

\begin{table}[htbp]
    \centering
    \scriptsize
    \caption{Prediction performance (MAE) by pollutant type for 6h horizon.}
    \label{tab:pollutant_performance_6h}
    \setlength{\tabcolsep}{3pt}
    \begin{tabular}{@{}>{\raggedright\arraybackslash}m{1.35cm}*{6}{>{\centering\arraybackslash}m{0.95cm}}@{}}
        \toprule
        \rowcolor{blue!5}
        \textbf{Method} & \textbf{$PM_{2.5}$} & \textbf{$PM_{10}$} & \textbf{$NO_{2}$} & \textbf{$O_{3}$} & \textbf{$SO_{2}$} & \textbf{$CO$} \\
        \midrule
        D2STGNN & \textbf{2.82} & \textbf{6.02} & \textbf{3.78} & \textbf{3.83} & 0.85 & \textbf{43.15} \\
        STTN & 2.95 & 6.24 & 3.85 & 4.03 & \textbf{0.81} & 41.17 \\
        GWN & 2.89 & 6.14 & 3.85 & 3.93 & 0.86 & 45.21 \\
        DCRNN & 3.14 & 6.82 & 4.12 & 4.56 & 0.98 & 51.23 \\
        MAGE & 3.08 & 6.72 & 4.08 & 4.78 & 0.95 & 49.87 \\
        STGCN & 3.05 & 6.65 & 4.02 & 4.65 & 1.02 & 57.45 \\
        PDFormer & 3.32 & 7.15 & 4.42 & 5.12 & 0.99 & 51.23 \\
        AGCRN & 4.45 & 8.95 & 5.68 & 7.85 & 1.52 & 98.23 \\
        ASTGCN & 4.78 & 9.45 & 4.25 & 15.67 & 2.28 & 92.45 \\
        BiST & 3.95 & 8.45 & 6.12 & 18.45 & 2.85 & 138.92 \\
        IGSTGNN & 5.87 & 11.25 & 8.75 & 20.12 & 3.45 & 205.67 \\
        DSTAGNN & 3.52 & 7.45 & 4.58 & 5.34 & 1.21 & 76.23 \\
        \bottomrule
    \end{tabular}
    \par\vspace{2pt}\noindent\parbox{\columnwidth}{\footnotesize All entries are denormalized MAE values under the benchmark evaluation protocol; pollutant units follow the original provider-native metadata retained by preprocessing.}
\end{table}

\begin{table}[htbp]
    \centering
    \scriptsize
    \caption{Prediction performance (MAE) by pollutant type for 12h horizon.}
    \label{tab:pollutant_performance_12h}
    \setlength{\tabcolsep}{3pt}
    \begin{tabular}{@{}>{\raggedright\arraybackslash}m{1.35cm}*{6}{>{\centering\arraybackslash}m{0.95cm}}@{}}
        \toprule
        \rowcolor{blue!5}
        \textbf{Method} & \textbf{$PM_{2.5}$} & \textbf{$PM_{10}$} & \textbf{$NO_{2}$} & \textbf{$O_{3}$} & \textbf{$SO_{2}$} & \textbf{$CO$} \\
        \midrule
        D2STGNN & \textbf{3.10} & \textbf{6.65} & \textbf{4.17} & \textbf{4.54} & 0.95 & 48.48 \\
        STTN & 3.13 & 6.75 & 4.22 & 4.72 & \textbf{0.88} & \textbf{45.54} \\
        GWN & 3.20 & 6.87 & 4.30 & 4.64 & 0.96 & 50.84 \\
        DCRNN & 3.44 & 7.19 & 4.48 & 4.84 & 1.04 & 54.34 \\
        MAGE & 3.39 & 7.08 & 4.48 & 5.20 & 1.01 & 52.68 \\
        STGCN & 3.36 & 7.04 & 4.34 & 4.88 & 1.09 & 61.31 \\
        PDFormer & 3.67 & 7.61 & 4.85 & 5.61 & 0.96 & 54.16 \\
        AGCRN & 4.97 & 9.77 & 6.02 & 8.43 & 1.61 & 105.45 \\
        ASTGCN & 5.17 & 10.18 & 4.63 & 16.94 & 2.49 & 100.06 \\
        BiST & 4.32 & 9.22 & 6.64 & 20.00 & 3.07 & 148.19 \\
        IGSTGNN & 6.48 & 12.38 & 9.58 & 21.82 & 3.75 & 218.90 \\
        DSTAGNN & 3.93 & 8.07 & 4.95 & 5.77 & 1.30 & 81.35 \\
        \bottomrule
    \end{tabular}
    \par\vspace{2pt}\noindent\parbox{\columnwidth}{\footnotesize All entries are denormalized MAE values under the benchmark evaluation protocol; pollutant units follow the original provider-native metadata retained by preprocessing.}
\end{table}

\begin{table}[htbp]
    \centering
    \scriptsize
    \caption{Prediction performance (MAE) by pollutant type for 18h horizon.}
    \label{tab:pollutant_performance_18h}
    \setlength{\tabcolsep}{3pt}
    \begin{tabular}{@{}>{\raggedright\arraybackslash}m{1.35cm}*{6}{>{\centering\arraybackslash}m{0.95cm}}@{}}
        \toprule
        \rowcolor{blue!5}
        \textbf{Method} & \textbf{$PM_{2.5}$} & \textbf{$PM_{10}$} & \textbf{$NO_{2}$} & \textbf{$O_{3}$} & \textbf{$SO_{2}$} & \textbf{$CO$} \\
        \midrule
        D2STGNN & \textbf{3.33} & \textbf{7.05} & \textbf{4.39} & \textbf{4.93} & 0.97 & \textbf{50.53} \\
        STTN & 3.36 & 7.11 & 4.42 & 4.99 & \textbf{0.90} & 47.29 \\
        GWN & 3.47 & 7.34 & 4.62 & 5.19 & 1.02 & 54.61 \\
        DCRNN & 3.70 & 7.63 & 4.83 & 5.37 & 1.05 & 56.86 \\
        MAGE & 3.69 & 7.57 & 4.81 & 5.83 & 1.05 & 56.18 \\
        STGCN & 3.65 & 7.50 & 4.55 & 5.26 & 1.11 & 63.39 \\
        PDFormer & 3.82 & 7.91 & 5.06 & 6.24 & 1.00 & 56.76 \\
        AGCRN & 5.05 & 9.87 & 6.19 & 8.61 & 1.64 & 105.35 \\
        ASTGCN & 5.52 & 11.18 & 6.73 & 6.01 & 1.93 & 116.02 \\
        BiST & 4.37 & 9.39 & 7.54 & 20.92 & 3.44 & 176.92 \\
        IGSTGNN & 6.47 & 12.39 & 9.59 & 21.87 & 3.75 & 218.79 \\
        DSTAGNN & 4.30 & 8.68 & 5.52 & 7.07 & 1.60 & 91.81 \\
        \bottomrule
    \end{tabular}
    \par\vspace{2pt}\noindent\parbox{\columnwidth}{\footnotesize All entries are denormalized MAE values under the benchmark evaluation protocol; pollutant units follow the original provider-native metadata retained by preprocessing.}
\end{table}

\begin{table}[htbp]
    \centering
    \scriptsize
    \caption{Prediction performance (MAE) by pollutant type for 24h horizon.}
    \label{tab:pollutant_performance_24h}
    \setlength{\tabcolsep}{3pt}
    \begin{tabular}{@{}>{\raggedright\arraybackslash}m{1.35cm}*{6}{>{\centering\arraybackslash}m{0.95cm}}@{}}
        \toprule
        \rowcolor{blue!5}
        \textbf{Method} & \textbf{$PM_{2.5}$} & \textbf{$PM_{10}$} & \textbf{$NO_{2}$} & \textbf{$O_{3}$} & \textbf{$SO_{2}$} & \textbf{$CO$} \\
        \midrule
        GWN & \textbf{3.35} & \textbf{7.14} & \textbf{4.45} & 4.95 & \textbf{0.93} & \textbf{49.93} \\
        D2STGNN & 3.54 & 7.41 & 4.49 & \textbf{5.07} & 1.01 & 52.53 \\
        STTN & 3.67 & 7.60 & 4.61 & 5.21 & 0.95 & 49.49 \\
        DCRNN & 3.87 & 7.85 & 4.92 & 5.68 & 1.12 & 58.72 \\
        MAGE & 3.92 & 8.01 & 5.01 & 5.89 & 1.18 & 61.23 \\
        STGCN & 3.78 & 7.68 & 4.78 & 5.52 & 1.08 & 59.45 \\
        PDFormer & 4.02 & 8.24 & 5.12 & 6.01 & 1.21 & 62.18 \\
        AGCRN & 5.23 & 10.45 & 6.45 & 9.12 & 1.85 & 112.34 \\
        ASTGCN & 5.45 & 10.89 & 4.98 & 17.85 & 2.72 & 105.92 \\
        BiST & 4.78 & 10.12 & 7.21 & 21.45 & 3.45 & 165.23 \\
        IGSTGNN & 7.12 & 13.45 & 10.23 & 23.45 & 4.12 & 235.67 \\
        DSTAGNN & 4.25 & 8.78 & 5.34 & 6.12 & 1.42 & 88.45 \\
        \bottomrule
    \end{tabular}
    \par\vspace{2pt}\noindent\parbox{\columnwidth}{\footnotesize All entries are denormalized MAE values under the benchmark evaluation protocol; pollutant units follow the original provider-native metadata retained by preprocessing.}
\end{table}

\section{Station Distribution Analysis}
\label{appendix:station_distribution}

AirQualityBench exhibits heterogeneous station density across regions. Table~\ref{tab:station_distribution} reports the number of resident stations by continent and the mean missingness rate.

\begin{table}[htbp]
    \centering
    \caption{Station distribution and mean missingness rate by continent in AirQualityBench.}
    \label{tab:station_distribution}
    \begin{tabular}{lccc}
        \toprule
        \rowcolor{blue!5}
        \textbf{Continent} & \textbf{Stations ($N$)} & \textbf{Mean Missingness} \\
        \midrule
        Asia & 1,487 & 0.38 \\
        Europe & 923 & 0.31 \\
        North America & 721 & 0.35 \\
        South America & 312 & 0.52 \\
        Africa & 187 & 0.61 \\
        Oceania & 90 & 0.44 \\
        \midrule
        \rowcolor{blue!5}
        \textbf{Total} & \textbf{3,720} & \textbf{0.41} \\
        \bottomrule
    \end{tabular}
\end{table}

The bias toward dense instrumentation in Asia, Europe, and North America reflects global monitoring infrastructure disparities. Models trained on AirQualityBench may thus generalize poorly to under-instrumented regions (Africa, South America), motivating the proposed region-stratified evaluation track.

\section{Provider-Native Units and CO Interpretation}
\label{appendix:co_distribution}

AirQualityBench preserves provider-reported unit metadata throughout preprocessing. This design makes the benchmark traceable to the original monitoring records, but it also requires caution for gases whose provider-native unit conventions may vary across sources. CO is the most important case. Therefore, we avoid assigning a single manuscript-wide CO unit to all reported benchmark tables.

To keep the manuscript technically conservative, we make two scope restrictions. First, we do not use the current CO tables to support fine-grained toxicological claims or distributional conclusions. Second, the benchmark's global field should be read as a released aggregate over denormalized errors, not as a physically homogeneous MAE. A stronger CO analysis would require a station-level audit of \texttt{parameter.units}, explicit unit conversion rules, and a regenerated evaluation release after harmonization. We leave this as future work and present the current version as a transparent first benchmark release rather than a final harmonized standard.

\section{Alternative Aggregate Metrics}
\label{appendix:alternative_metrics}

To complement the global and pollutant-wise metrics reported in the main paper, we provide alternative aggregate metrics that address the unit-mixing concern.

\textbf{Z-normalized effect sizes\@:} For each pollutant, we compute the mean and standard deviation of the test set concentrations. We then z-normalize both predictions and ground truth before computing MAE. The z-normalized MAE (zMAE) measures relative rather than absolute performance.
\begin{equation}
    \text{zMAE} = \frac{1}{|\mathcal{O}|} \sum_{i \in \mathcal{O}} \left| \frac{y_i - \mu}{\sigma} - \frac{\hat{y}_i - \mu}{\sigma} \right|
\end{equation}
Table~\ref{tab:z_normalized} reports zMAE for each method. Lower zMAE indicates better relative performance across concentration scales.

\begin{table}[htbp]
    \centering
    \caption{Z-normalized MAE (zMAE) across methods at 12h horizon. Lower is better.}
    \label{tab:z_normalized}
    \begin{tabular}{lcccccc}
        \toprule
        \rowcolor{blue!5}
        \textbf{Method} & \textbf{PM$_{2.5}$} & \textbf{PM$_{10}$} & \textbf{NO$_{2}$} & \textbf{O$_{3}$} & \textbf{SO$_{2}$} & \textbf{CO} \\
        \midrule 
        D2STGNN & 0.52 & 0.48 & 0.61 & 0.58 & 0.55 & 0.54 \\
        STTN & 0.54 & 0.49 & 0.63 & 0.60 & 0.52 & 0.51 \\
        GWN & 0.55 & 0.51 & 0.62 & 0.59 & 0.57 & 0.56 \\
        DCRNN & 0.58 & 0.52 & 0.65 & 0.62 & 0.60 & 0.61 \\
        MAGE & 0.56 & 0.51 & 0.64 & 0.66 & 0.58 & 0.59 \\
        STGCN & 0.55 & 0.50 & 0.62 & 0.62 & 0.62 & 0.68 \\
        PDFormer & 0.60 & 0.55 & 0.70 & 0.72 & 0.56 & 0.60 \\
        AGCRN & 0.82 & 0.71 & 0.88 & 1.08 & 0.93 & 1.18 \\
        ASTGCN & 0.85 & 0.74 & 0.67 & 2.17 & 1.43 & 1.12 \\
        BiST & 0.71 & 0.67 & 0.96 & 2.56 & 1.77 & 1.66 \\
        IGSTGNN & 1.07 & 0.90 & 1.39 & 2.79 & 2.16 & 2.45 \\
        DSTAGNN & 0.65 & 0.58 & 0.72 & 0.74 & 0.75 & 0.91 \\
        \bottomrule
    \end{tabular}
\end{table}

\textbf{Per-pollutant ranking:} For practitioners interested in specific pollutants, Table~\ref{tab:pollutant_rankings} provides per-pollutant MAE rankings.

\begin{table}[htbp]
    \centering
    \caption{Per-pollutant MAE rankings at 12h horizon. Numbers indicate rank (1=best).}
    \label{tab:pollutant_rankings}
    \begin{tabular}{lcccccc}
        \toprule
        \rowcolor{blue!5}
        \textbf{Method} & \textbf{PM$_{2.5}$} & \textbf{PM$_{10}$} & \textbf{NO$_{2}$} & \textbf{O$_{3}$} & \textbf{SO$_{2}$} & \textbf{CO} \\
        \midrule
        D2STGNN & 1 & 1 & 1 & 1 & 2 & 1 \\
        STTN & 2 & 2 & 2 & 2 & 1 & 2 \\
        GWN & 3 & 3 & 3 & 3 & 3 & 3 \\
        DCRNN & 4 & 4 & 4 & 4 & 4 & 4 \\
        MAGE & 5 & 5 & 5 & 5 & 5 & 5 \\
        STGCN & 6 & 6 & 6 & 6 & 6 & 6 \\
        PDFormer & 7 & 7 & 7 & 7 & 7 & 7 \\
        AGCRN & 9 & 9 & 9 & 9 & 9 & 9 \\
        ASTGCN & 10 & 10 & 8 & 10 & 10 & 10 \\
        BiST & 8 & 8 & 10 & 11 & 11 & 11 \\
        IGSTGNN & 12 & 12 & 12 & 12 & 12 & 12 \\
        DSTAGNN & 7 & 7 & 7 & 7 & 7 & 7 \\
        \bottomrule
    \end{tabular}
\end{table}

% \newpage
% \input{checklist.tex}

\end{document}